\documentclass[preprint, authoryear, 11pt]{elsarticle}
\usepackage[utf8]{inputenc}
\usepackage[T1]{fontenc}
\usepackage[default,oldstyle, scale=.95]{opensans} 
\usepackage{hyperref}
\usepackage{url}
\usepackage[a4paper, margin=1in]{geometry}
\usepackage{microtype}
\usepackage{titlesec}
\usepackage[dvipsnames]{xcolor}
\usepackage{mathpazo}
\usepackage{tocbibind}
\usepackage{graphicx,wrapfig,lipsum}
\usepackage{eurosym}
\usepackage{pgfplots}
\usepgfplotslibrary{groupplots}
\pgfplotsset{compat=1.18}
\usepackage{tikz}
\usepackage{amsmath,amssymb,amsfonts}
\usepackage{algorithmic}
\usepackage{textcomp}
\usepackage{multirow}
\usepackage{float}
\usepackage{makecell}
\usepackage{subcaption}
\usepackage{cleveref}
\usepackage{siunitx}
\usepackage{natbib}
\usepackage{setspace}

\titleformat{\section}
  {\color{blue}\normalfont\Large\bfseries}
  {\thesection}{1em}{}

\titleformat{\subsection}
  {\color{purple}\normalfont\large\bfseries}
  {\thesubsection}{1em}{}

\titleformat{\subsubsection}
  {\color{teal}\normalfont\normalsize\bfseries}
  {\thesubsubsection}{1em}{}

\titleformat{\paragraph}[block]{\color{ForestGreen}\normalfont\normalsize\bfseries}{\theparagraph}{1em}{}
\titlespacing*{\paragraph}{0pt}{1.25ex plus 1ex minus .1ex}{1em}

\makeatletter
\@addtoreset{paragraph}{subsubsection} 
\makeatother

\journal{Transportation Research Part - C}

\definecolor{brightblue}{rgb}{0.0, 0.5, 1.0}

\hypersetup{
    pdftitle={Deep Learning-Based Mesoscopic Emission Modeling},
    pdfauthor={DIB Abdelkader, BALAC Milos, SCIARRETTA Antonio},
    pdfsubject={Research Paper},
    pdfkeywords={Machine learning, Deep learning, Traffic emissions, Transportation modeling, Emission modeling, Urban air pollution},
    colorlinks=true,
    linkcolor=cyan,
    filecolor=cyan,      
    urlcolor=cyan,    
    citecolor=cyan,  
    pdfpagemode=FullScreen
}

\begin{document}
\setstretch{1.0}

\begin{frontmatter}

\title{Deep Learning-Based Mesoscopic Emission Modeling}

\author[a]{DIB Abdelkader\corref{cor1}}
\cortext[cor1]{Corresponding author}
\ead{dabdelkader@ethz.ch}

\author[a]{BALAC Milos}
\ead{balacm@ethz.ch}

\author[b]{SCIARRETTA Antonio}
\ead{antonio.sciarretta@ifpen.fr}

\address[a]{Center for Sustainable Future Mobility, ETHZ, Zurich, Switzerland}
\address[b]{IFP Energies nouvelles, Rueil-Malmaison, France}

\begin{abstract}


{
Traffic-related emissions are a major contributor to urban air pollution. However, existing emission models face a fundamental trade-off: macroscopic models are computationally efficient but rely on average speeds, leading to systematic errors, while microscopic models achieve high accuracy but are computationally prohibitive for large scale simulations. To bridge this gap, this study introduces Direct Emission Prediction (DEP), a Gated Recurrent Unit (GRU)-based deep learning model. DEP was trained on over one million synthetic trips generated by coupling real-world GNSS speed profiles with the microscopic emission model MOBSim. By processing sequential link-level inputs, including vehicle characteristics, road attributes, and ambient temperature, DEP predicts CO\textsubscript{2} and NOx emissions. We integrated DEP into a MATSim-based workflow and applied it to the Île-de-France region. DEP matches the accuracy of SPG-M, a hybrid mesoscopic–microscopic reference model, within 0.35\% for CO\textsubscript{2} and 1.87\% for NOx across both hot and cold-start scenarios, while reducing computational runtime to just 2.2\%. Furthermore, validation against empirical remote sensing data confirms that DEP accurately reproduces real-world NOx emission levels across Euro 2–6 diesel vehicles, whereas traditional macroscopic tools like COPERT and HBEFA systematically underestimate them. Finally, DEP effectively captures the spatial distribution of cold-start emissions, identifying key hotspots in dense urban areas. This work demonstrates that deep learning-based mesoscopic frameworks can achieve microscopic accuracy at a fraction of the computational cost, enabling scalable, high-resolution urban emission modeling without relying on pre-computed emission factor databases.
}
\end{abstract}
\end{frontmatter}

\section{Introduction}
{
Outdoor air pollution is one of the most pressing public health challenges of our time. It causes at least 4.2 million deaths annually worldwide according to the \citet{who_air_pollution}, and more than \num{350000} deaths in the European Union (EU) in 2022 alone \citep{eea_air_pollution_2024}. When indoor air pollution is included, the two together account for approximately 10\% of all deaths worldwide, making air pollution the fourth leading mortality risk globally \citep{shaw2022evaluation}. Beyond mortality, the consequences extend across multiple dimensions of human wellbeing: exposure to particulate matter (PM) and nitrogen oxides (NOx) impairs cognitive function and is associated with mental health disorders, sleep disturbances, anxiety, depression, and suicide \citep{chang2019cognitiveimpact, ebenstein2016cognitiveimpact}. The economic toll is equally significant, a 10\% reduction in fine particulate pollution is estimated to increase EU economic output by between 0.5\% and 1.5\%, while the benefits of the U.S. Clean Air Act Amendments over the period 1990--2020 are estimated at 12 trillion USD \citep{dechezlepretre2019economicimpact}.}

{
Road traffic is a major driver of this burden, particularly in urban areas. In Paris, road traffic alone is estimated to account for 30\% of greenhouse gas emissions and 56\% of NOx emissions according to \citet{AirparifParis}. Accurate estimation of traffic-related emissions is therefore essential for developing effective mitigation strategies and informing policy decisions. To this end, various modeling approaches have been developed, primarily categorized as top-down or bottom-up methods \citep{pinto2020trafficreview, zhang2019review}.}

Top-down models rely on aggregate data and statistical techniques to estimate emissions at large spatial scales. They correlate historical emission levels with macro-level variables such as fuel consumption, population trends, economic indicators, and vehicle fleet composition \citep{MC2, pinto2020trafficreview}. While suitable for national or regional inventories, their limited spatial resolution makes them unsuitable for evaluating localized pollution patterns or targeted interventions.

In contrast, bottom-up methods utilize fine-grained data to offer more accurate emission estimates across multiple dimensions \citep{zhang2019review}. This enhanced capability is vital for a comprehensive understanding of the environmental impact of traffic in specific areas or districts, and facilitates precise modeling of diverse traffic scenarios \citep{piccoli2024bottom, spgm}. However, their implementation at larger scales presents significant challenges.

Developing a bottom-up model to estimate road traffic pollutant emissions with high spatial and temporal resolution, as well as prospective capabilities, often involves the use of traffic simulations \citep{mkadziel2023vehicle}. Traffic simulations model the dynamic interactions between populations and transportation systems \citep{dorokhin2020traffic}. These simulations produce data that are integrated into emission models to estimate pollutant levels across both space and time. However, there is a trade-off between the level of detail provided by these models and their computational efficiency.

For instance, the computational demands of microscopic traffic simulations paired with microscopic emission models, while providing highly detailed estimates for various kinds of studies \citep{MISRA201381, vissim_vsp, mkadziel2023investigation}, effectively limit their application to small geographic areas such as individual intersections or specific traffic corridors \citep{tu2018development}. Conversely, mesoscopic and macroscopic approaches offer greater computational efficiency for large-scale networks, though less detailed. For example, Airparif uses a macroscopic traffic model combined with COPERT (COmputer Programme to calculate Emissions from Road Transport, \citet{copert}) to estimate emissions in its Île-de-France inventory \citep{Airparif2019}. Similar methods have been applied in cities like Madrid \citep{borge2012comparisonMadrid}, Barcelona \citep{rodriguez2021coupledBarcelona}, Paris \citep{elessa2018olympus}, and New York City \citep{macroHbefa}, where macroscopic traffic models are paired with emission models to estimate road traffic emissions. In Florida, MATSim (Multi-Agent Transport Simulation, \citet{matsim}), which is an agent-based traffic simulation model classified as mesoscopic due to its use of a queue-based approach to represent traffic dynamics, was combined with a macroscopic emission model to evaluate individual exposure to air pollution and assess related inequalities \citep{gurram2019matsimMoves}.

Although macroscopic models are computationally efficient for large-scale simulations, they often produce less accurate emission estimates \citep{rodriguez2021coupledBarcelona, tu2018development}, particularly when trying to capture local emission peaks. For instance, comparisons between COPERT and HBEFA (Handbook Emission Factors for Road Transport, \citet{hbefa}) models combined with macroscopic traffic models revealed significant differences in outputs \citep{borge2012comparisonMadrid}. Similarly, in \citet{grabe2022}, a model based on MATSim combined with HBEFA underestimated emissions compared to real-world Portable Emission Measurement Systems (PEMS) measurements. In addition, multiple studies like \citet{rodriguez2021coupledBarcelona,tu2018development}, and \citet{underEstimationNOx1} highlight discrepancies between macroscopic and microscopic approaches, especially in urban areas with heavy traffic. Even when macroscopic models provide reasonable aggregate estimates, they struggle to capture the spatial distribution of emissions \citep{gecoAiremissions, mohAndGiovaniUsingCopert}. {This is largely due to the fact that macroscopic emission models primarily rely on either average speed or discrete traffic situations, neither of which can always accurately reflect actual driving dynamics. Traffic-situation models, such as HBEFA, face challenges due to the level of data aggregation and the inherent ambiguity in defining traffic conditions \citep{theseEmissionsLondon, samaras2019improvedCopert}. For average-speed-based models, such as COPERT, it has been shown that relying on average speed alone, without accounting for congestion or acceleration patterns, can lead to significant inaccuracies in emission estimates \citep{samaras2019improvedCopert, smit2008air}. For instance, a speed of 45 km/h may represent free-flow conditions in urban areas, yet indicate heavy congestion on highways, resulting in fundamentally different emission profiles.}

This limitation can lead to misguided policy decisions, such as implementing measures in incorrect locations or neglecting critical areas. Additionally, the impact of mitigation measures, including their location, extent, and direction (whether emissions increase or decrease), may differ from expectations \citep{wismans2013comparison}.

These limitations have led researchers to explore alternative approaches to improve accuracy. Hybrid methods, combining microscopic and macroscopic models with machine learning, have been developed in some studies. For example, in \citet{tu2018development}, a hybrid approach applied microscopic traffic and emission models to a sample of network links. These links were categorized using unsupervised learning based on traffic conditions, and representative emission factors were assigned to each category. The categorized data and emission factors were then integrated into a mesoscopic traffic simulation for large-scale analysis. In \citet{spgm}, a new workflow named SPG-M (Speed Profile Generation -- MOBSim) was introduced, combining the mesoscopic traffic model MATSim with a microscopic emission model. This was accomplished by generating high-resolution speed profiles using deep learning. Specifically, MATSim outputs were converted into instantaneous speed profiles using a deep learning generator (SPGen) developed in \citet{spgen}, allowing for the integration of a microscopic emission model into the simulation workflow.

Alternative data-driven methods have also been proposed in the literature. Data-driven models for traffic flow estimation are increasingly prevalent in the literature \citep{giov1, giov2}, leveraging GPS signals, traffic sensors, and speed probes to predict traffic dynamics. These models are typically combined with macroscopic emission models to estimate emissions \citep{dataDrivenEmissions, dataDrivenEmissions2}. End-to-end models predicting pollutant concentrations directly from traffic data also exist \citep{DU2023745,yang2021covid}. However, these methods face limitations in analyzing diverse prospective scenarios due to omitted critical parameters and challenges in generalizing to new conditions outside the training data. This limitation is particularly significant in prospective analyses where relationships between variables may differ from historical patterns, reducing the adaptability of machine learning models.

This research introduces a new methodology that combines a mesoscopic traffic model with a mesoscopic emission model. The approach builds on the SPG-M framework, which integrates the same traffic model with a microscopic emission model through deep learning-based speed profile generation. However, the use of a microscopic emission model was found to be computationally expensive for large-scale applications, resulting in reduced workflow efficiency. To address this limitation, we replaced both the speed profile generation step and the microscopic emission model with a highly efficient deep learning-based mesoscopic emission model called DEP (Direct Emission Prediction). An initial version of this emission model was introduced in \citet{emgen} and has been further refined and validated in this study.

The emission model is trained on a large synthetic dataset of emissions. The goal of this approach is to maintain the same level of accuracy as in SPG-M, both globally and locally, while significantly improving computational efficiency.

{The proposed framework makes three key contributions to traffic emission modeling:}
\begin{itemize}
    \item {Unlike macroscopic models, DEP processes the full sequence of links traversed by each vehicle, capturing speed transitions and the resulting  acceleration patterns that drive local emission peaks, a capability previously limited to microscopic approaches, while remaining computationally efficient enough for region-scale deployment.}
    \item {Whereas conventional macroscopic models provide no spatial distribution of cold-start emissions, DEP explicitly tracks each vehicle throughout its entire trip, enabling the attribution of cold-start excess emissions to specific links and urban zones.}
    \item {By replacing the computationally intensive speed profile generation and microscopic emission estimation steps of SPG-M with a single deep learning model, DEP reduces runtime to 2.2\% of the SPG-M simulation time while maintaining comparable accuracy at both global and local scales.}
\end{itemize}

Section \ref{sec:methodology} outlines the methodology, detailing the workflow, its components, and their validation. Section \ref{sec:results} presents the findings and compares them with existing measures and macroscopic models. Section \ref{sec:conclusion} concludes the paper, summarizing the key insights.

\section{Methodology}\label{sec:methodology}
The emission estimation workflow is structured into two primary components: traffic simulation and emission modeling. This study employs the open Île-de-France (IdF) scenario \citep{matIdf3, matIdf1} for traffic simulation using MATSim. Emissions are then estimated using DEP.

\begin{figure}[h]
    \centering
    \includegraphics{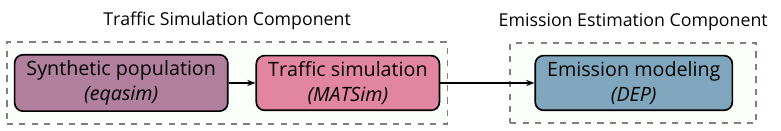}
    \caption{General workflow of DEP for traffic emission modeling}
    \label{fig:workflow}
\end{figure}

The workflow structure is shown in \autoref{fig:workflow}. First, raw datasets are processed using the eqasim pipeline \citep{eqasim} to generate a synthetic population that represents realistic travel demand. This population is then used as input for MATSim simulations. After the simulation, all car trips are collected. For each trip, the vehicle is tracked along its route from origin to destination to record its average speed on each link it traversed. Finally, DEP uses these average speeds, together with road and vehicle attributes, to estimate NOx and CO\textsubscript{2} emissions for each trip and within each link.

\subsection{Synthetic Population and Travel Demand}
A synthetic population refers to a set of artificially generated agents that statistically represent the real population of a specific geographic region. Each agent corresponds to an individual and is characterized by a set of socio-demographic attributes, such as age, gender, and income, that influence decision-making processes within simulations.

In the context of travel demand modeling with MATSim, each agent is also assigned a daily activity schedule, specifying the timing and spatial location at the coordinate level of various activities. These activities are connected by trips that the agent is expected to undertake to fulfill their daily plan. MATSim then simulates these trips dynamically within the transportation network, capturing temporal and spatial interactions among agents.

To enable the simulation of traffic-related emissions, it is essential that agents who own vehicles in reality are assigned corresponding vehicles in the synthetic population. Since emissions are influenced by vehicle characteristics, including fuel type and emissions standard, these attributes must also be generated and assigned to each synthetic vehicle. This allows for a more accurate estimation of emissions at both the individual and global levels.

The methodology employed to create the synthetic population for the Île-de-France (IdF) region is detailed in our previous work \citep{spgm}. In this section, we provide a concise overview of the main components of that approach.

\subsubsection{Synthetic Population Generation}
The model employs the eqasim pipeline, an open-source Python tool that integrates publicly available datasets to generate a realistic synthetic population. Each agent in this population has detailed sociodemographic attributes and a plan of daily activities scheduled in both time and space. The framework also constructs the road network using OpenStreetMap (OSM) data and generates public transit schedules from GTFS data provided by Île-de-France Mobilités.

\subsubsection{Cross-Border Traffic Modeling}
To address the limitation that eqasim only includes trips fully contained within the Idf region, we generated an external population representative of the cross-border traffic, based on cordon surveys conducted between 2013 and 2016 \citep{DRIEAT2016}. These surveys, capturing approximately 90\% of inbound and outbound traffic at 20 primary road network stations, allowed us to create external agents matched to recorded traffic volumes. Activity locations and types were assigned based on survey data, with sociodemographic attributes generated to reflect the overall regional population distribution in the Idf region.

{To our knowledge, these surveys represent the most recent publicly available data on cross-border traffic in the Île-de-France region. Although traffic patterns may have changed over the past decade, we believe it is preferable to include this data, despite its age, rather than omit cross-border flows entirely, as the latter option would result in a systematic underestimation of traffic volumes at the network boundaries and on the highways.}

\subsubsection{Road Network Implementation}
While the eqasim pipeline primarily utilizes OSM data, our model incorporates an additional stage allowing selection between OSM and HERE \citep{here} networks. The HERE network was primarily adopted for compatibility with subsequent stages in the workflow, particularly with the training database of DEP. Beyond this compatibility benefit, the HERE network provides enhanced detail, including speed limits, lane counts, slope data, and traffic infrastructure information, attributes valuable for emission modeling.

\subsubsection{Vehicle Fleet Modeling}
The vehicle generation and assignment methodologies in the eqasim pipeline have been enhanced. The original implementation assigned vehicles to agents randomly based on their commune of residence and did not include detailed vehicle attributes. Two key improvements were introduced to address these limitations.

First, the updated pipeline now generates essential vehicle attributes such as weight and power, which are necessary for microscopic emission modeling but not required by macroscopic models. {These attributes are derived from French vehicle fleet data as of 1 January 2023 \citep{FranceAuto2023} and ADEME car labeling data covering vehicles registered up to 2022 \citep{CarLabellingADEME}, which provide comprehensive environmental information on private vehicles sold in France.}

Second, the vehicle assignment algorithm has been refined to account for agents' socioeconomic characteristics. Rather than a purely random commune-based assignment, the model now incorporates agent income when allocating vehicles. The higher the agent's income, the higher the probability of possessing a less polluting car. These probability distributions are derived from the mobility survey data \citep{MobilitySurvey2019}, resulting in a more realistic representation of vehicle ownership patterns across different income segments.

\subsection{Traffic Simulation}

Our traffic simulation framework is based on MATSim, an open-source, agent-based platform designed for modeling large-scale transportation systems \citep{matsim}. The model employs a queue-based traffic dynamics approach, which strikes a balance between computational efficiency and realistic representation of traffic behavior, making it particularly suitable for large-scale applications. In this framework, each road link is modeled as a queue with defined storage capacity, flow capacity, and free-flow speed, allowing for the simulation of congestion and spillback effects.

MATSim operates using an iterative co-evolutionary algorithm. In each iteration, agents simulate their daily activities and travel patterns. A small subset of agents (5\% in this study) then improves their plans based on predefined strategies. This process repeats until the system converges to a state of equilibrium.

For our IdF implementation, the model incorporates five transportation modes: walking, cycling, driving, car passenger, and public transport. Mode choice behavior is modeled using the Discrete Mode Choice extension \citep{dcm1,dcm2}. To improve computational efficiency, a common practice in large-scale agent-based simulations is to use a sample of the population \citep{llorca2019effects}. In this study, we use a 5\% sample of the total population.

Detailed descriptions of the model implementation, calibration methodology, and validation procedures can be found in \citet{spgm}.

\subsection{Emission Model}
This section presents the design and development of the deep learning-based mesoscopic emission model, DEP, used to estimate link-level pollutant emissions. It outlines the methodology underlying data generation, preprocessing strategies, and the architecture, training, and validation of the model.

\subsubsection{Data Preparation}
Deep learning models rely heavily on datasets that capture a wide variety of conditions and scenarios for their development. Collecting such large datasets is expensive, especially when they must be sourced from real-world, on-road experiments. The challenge grows when data must account for diverse driving conditions, vehicle types, and routes.

To address this, this study uses a practical and cost-effective method for gathering data to train the model. It combines real-world speed profiles obtained from a GNSS database with the microscopic emission model, MOBSim \citep{mobsim2, gecoAiremissions}. The speed profile dataset, previously used to train a speed profile generator, is detailed in \citet{spgen} and originates from the GecoAir crowdsensing application \citep{gecoair}.

{
It is worth emphasizing that this methodology does not rely on emission factors in the classical sense. Rather than associating a vehicle category or speed bin with a pre-aggregated emission value, DEP directly predicts link-level emissions for each individual vehicle given its full trip context. MOBSim serves as the reference emission model to prepare the training data, it is calibrated to the French vehicle fleet and used both to generate the training dataset for DEP and as the reference in the SPG-M workflow. The accuracy of DEP is therefore inherently bounded by that of MOBSim.
}

This approach is widely employed in developing data-driven emission models. For instance, speed profiles are derived from a microscopic traffic model, and a microscopic emission model is used in \citet{samaras2019improvedCopert} to create the dataset. This method reduces costs while providing a diverse dataset to simulate various driving conditions effectively.

\paragraph{Data Generation}
The dataset of speed profiles comprises {more than $440 \thinspace 000$} individual speed profiles. Each speed profile represents the vehicle’s speed measurements recorded from the origin to the destination of a trip, sampled at 1 Hz (one speed value per second). Each speed value is associated with a corresponding link in the HERE network, enabling the retrieval of related infrastructure data.

These profiles are geographically distributed across the IdF and Lyon metropolitan regions in France and represent a diverse range of driving conditions. The trips were conducted by more than $5 \thinspace 800$ distinct drivers.

This study also uses a vehicle dataset obtained from the same source as the speed profiles \citep{gecoair}, containing over $11 \thinspace 000$ vehicles with detailed attributes.

The data generation procedure is designed to simulate a variety of realistic driving scenarios. It starts by randomly selecting a speed profile (The selected data includes road gradient), then randomly assigns a vehicle from the vehicle dataset to this profile. The process also randomly sets the initial engine temperature between 0°C and 90°C, representing various engine start conditions. Additionally, the ambient temperature is randomly chosen from a range between 0°C and 40°C to depict different climatic conditions, with the constraint that it must not exceed the initial engine temperature.

The simulations are then conducted using MOBSim. This simulation records data related to the vehicle’s performance and emissions on a second-by-second basis. Subsequent data processing involves aggregating the instantaneous emissions data by link. 

{This approach is used to produce a comprehensive dataset of over 1 million simulated trips, as each speed profile is paired with multiple randomly selected vehicles and initial temperature conditions.}

The dataset encompasses diverse features with significantly different scales and units: vehicle speeds (km/h), NOx emissions (mg), CO\textsubscript{2} emissions (g), temperatures (°C), vehicle weights (kg), power (W), and binary categorical variables. To prevent features with larger numerical ranges from dominating the model training and to ensure balanced learning, data scaling is necessary.

\paragraph{Data Scaling}

The application of appropriate data scaling techniques represents a fundamental preprocessing step that substantially influences machine learning model performance \citep{ahsan2021effect, sujon2024use}. The selection of an optimal scaling technique depends on both the machine learning model's architecture and the dataset's intrinsic properties. 

A particularly challenging scenario is encountered in the scaling of the created dataset containing NOx emissions. This dataset often reflects a substantial variance in emission levels between older diesel vehicles and those running on gasoline or equipped with newer aftertreatment technologies. If separate models were developed for each vehicle category, based on similar emission levels, this variance might not pose a significant issue. However, the development of a unified model designed to predict NOx emissions across all vehicle types necessitates a uniform scaling approach to ensure that all emissions data fall within the same scale.

To achieve this, each vehicle is assigned a scaling factor based on its fuel type and Euro norm. This factor is then applied to adjust the NOx emissions data, bringing these values for different vehicle types to a comparable scale.

\begin{figure}[h]
    \centering
    \includegraphics[width=0.8\textwidth]{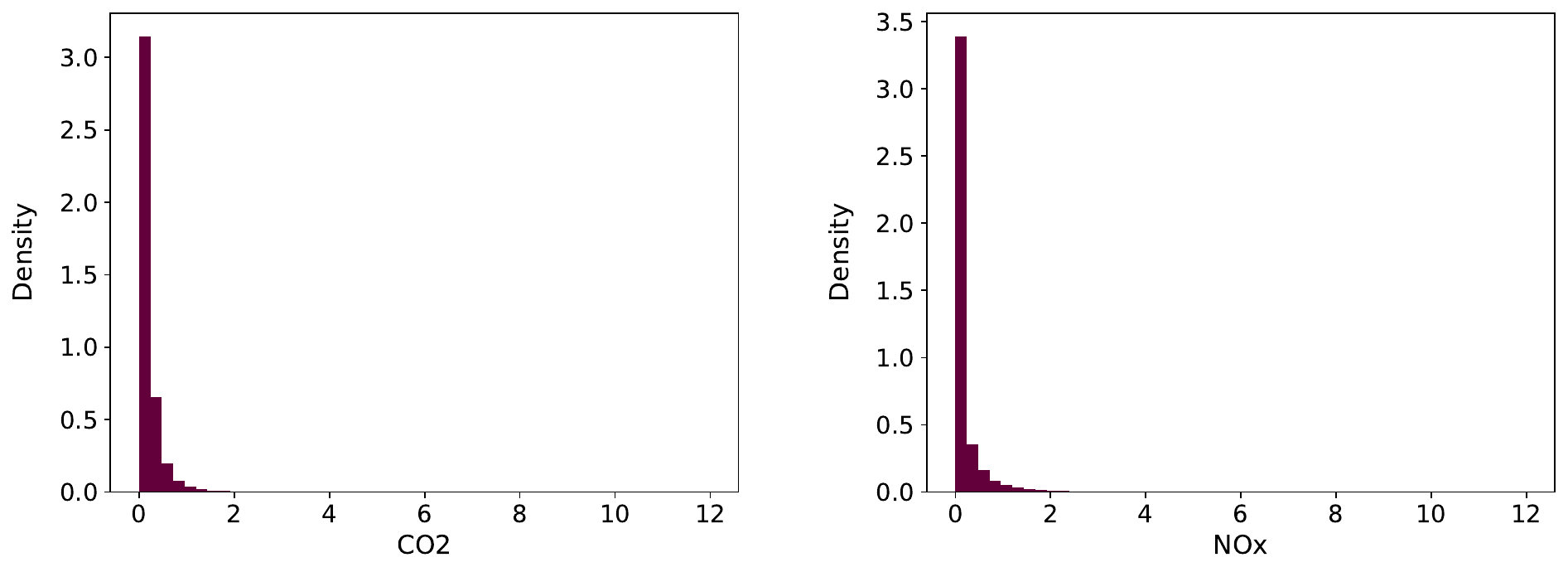}
    \caption{Histograms of scaled CO\textsubscript{2}  and NOx emissions in the training dataset}
    \label{fig:emissionsScale}
\end{figure}

Even for the same vehicle, emission levels can vary significantly from one link to another, influenced by factors such as acceleration, speed fluctuations, road gradient, and link length. Emissions exhibit a right-skewed distribution, where lower values are more common and very high values occur less frequently but are important to capture accurately. While transformations such as the Box-Cox, Yeo-Johnson, or logarithmic transformation may initially appear advantageous due to their ability to approximate a normal distribution \citep{Albada2007Transformation,Raymaekers2020Transforming}, they introduce notable challenges. Despite enhancing model stability during training, these nonlinear transformations lead to significant errors during inference when applying the inverse transformation. This issue arises because small errors in the transformed space are amplified upon reversing the transformation, resulting in considerable discrepancies in the original, untransformed space. Therefore, emission data are scaled by the interquartile range. The resulting distributions of CO\textsubscript{2} and NOx data are presented in \autoref{fig:emissionsScale}.

\begin{figure}[h]
    \centering
    \includegraphics[width=0.95\textwidth]{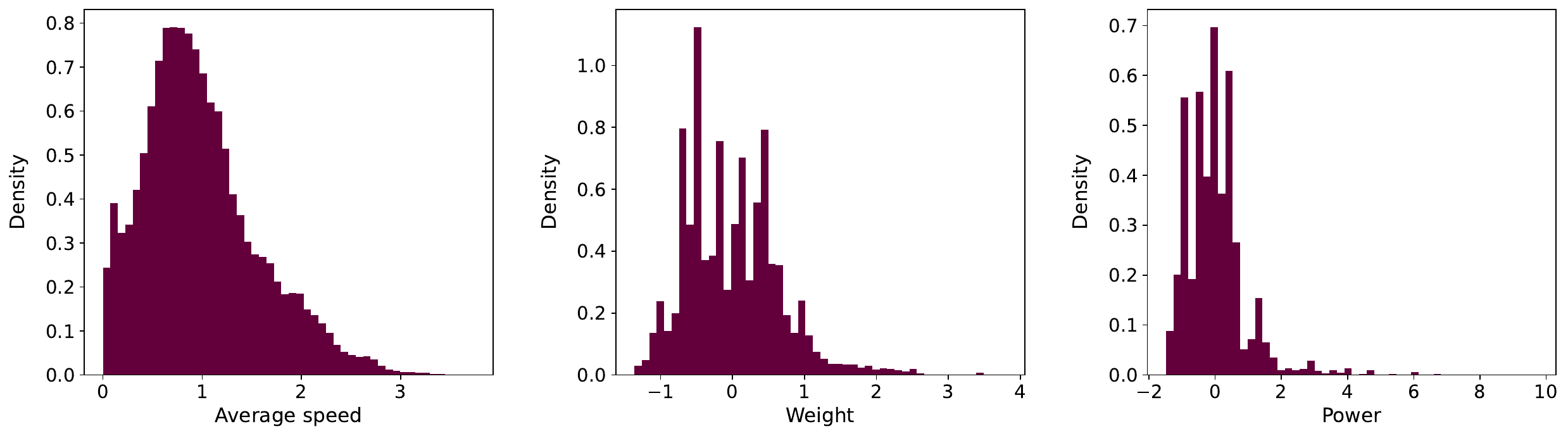}
    \caption{Histograms of scaled average speed, vehicle weight, and engine power in the training dataset}
    \label{fig:speedWeightPowerDistributions}
\end{figure}

Weight, power, ambient temperature, and initial engine temperature are standardized using z-scores. {Link length, however, exhibits high variability, ranging from 8 meters to over 1000 meters, with a right-skewed distribution concentrated below 200 meters. Several scaling strategies were evaluated, including Box-Cox transformation; however, this led to overestimation of emissions, as it compressed the wide range of link lengths into a narrower scale, preventing the model from capturing emission variations associated with different link lengths. As is common in data-driven modeling, the choice of scaling method depends on the specific data properties and model architecture, and is ultimately guided by empirical performance \citep{sujon2024use}. A simple division by 100 proved most effective, preserving the dynamic range while keeping values within a numerically suitable scale.} Similarly, velocities are scaled by their average, which is 40 km/h. The distributions of weight, power, and average speed per link are illustrated in \autoref{fig:speedWeightPowerDistributions}.

\subsubsection{Architecture}
{Vehicle emission levels are determined by a complex combination of factors, some of which are not always easily accessible for precise evaluation. These factors include driving behavior (such as speed and acceleration patterns), vehicle-specific attributes (such as engine specifications, weight, and exhaust aftertreatment technology type), road conditions (such as slope), and weather conditions \citep{mobsim4, mobsim1,ref1reviewer}.}

It is important to recognize that these influencing factors do not have a uniform impact on both pollutant emissions and vehicle fuel consumption. As a result, considering all these parameters in a thorough manner may not be practical due to the inherent challenges in data collection and the complexity of developing a comprehensive emission model. In addition, incorporating highly detailed attributes does not necessarily improve accuracy and can even reduce performance when such data is unavailable or of poor quality, as demonstrated in \cite{SMIT20102943}.

Therefore, this study adopts a practical approach by focusing on a subset of key parameters for estimating emissions at the link level. Vehicle-related inputs include weight, engine power, Euro norm, vehicle category, and fuel type. Trip-related inputs comprise link length, average speed, road slope, and speed limit, as the combination of average speed and speed limit enables the estimation of congestion levels. To account for cold-start emissions and environmental conditions, the initial engine temperature, initial temperature of exhaust aftertreatment devices, and ambient temperature are also included.

{The initial temperature of exhaust aftertreatment devices is a critical parameter for NOx emission estimation, as these systems only reach full efficiency once sufficiently warmed up \citep{ref1reviewer, ref2reviewer, ref3reviewer}. By explicitly including this temperature as an input, DEP is able to track the thermal state of aftertreatment systems throughout the trip and capture the excess NOx emissions associated with cold or partially warmed devices.}

By restricting the scope to these selected parameters, the aim is to balance the practicality of data collection with the goal of providing meaningful and reasonably accurate estimates of pollutant emissions. 

\begin{figure}[!t]
    \centering
    \includegraphics[angle=-90, width=0.6\textwidth]{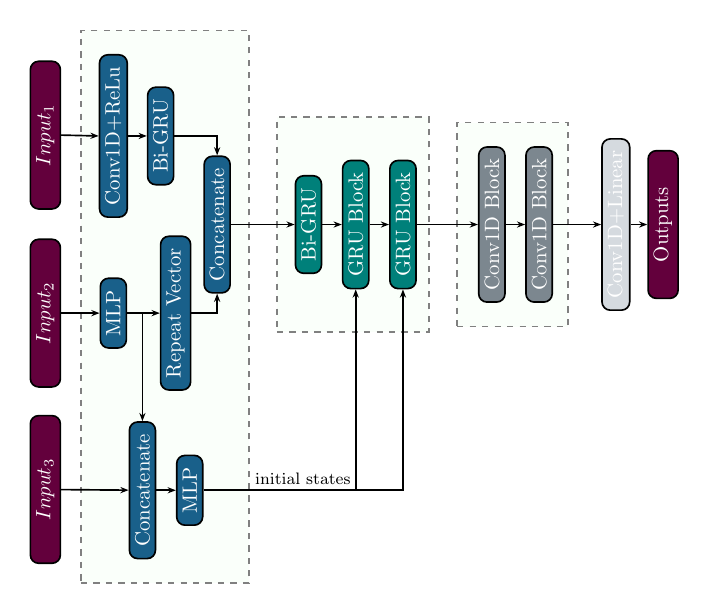}    
    \caption{DEP emission model architecture}
    \label{fig:model-architecture}
\end{figure}

The model architecture, illustrated in \autoref{fig:model-architecture}, is inspired by the first stage of SPGen, which successfully predicted speed dynamics. By adding additional layers, SPGen first stage could be adapted to predict emissions. However, instead of using Long Short-Term Memory (LSTM; \citet{lstm}) networks as in SPGen, this model employs Gated Recurrent Unit (GRU; \citet{gru}) networks to enhance efficiency, reducing the number of parameters and making the model lightweight. Unlike many mesoscopic and macroscopic emission models that are not sequential, the decision to use a sequential model is based on the need  to account for speed transitions between links, enabling the model to capture variations in speed and acceleration, and thus better represent local emissions. {To illustrate, consider a vehicle traveling at 70 km/h on a given link ($l_i$). This observation alone is consistent with three fundamentally different driving situations: the vehicle may have accelerated from 20 km/h on the previous link, cruised at a constant 70 km/h over a long distance, or decelerated from 130 km/h. Each scenario corresponds to a markedly different emission profile, yet a non-sequential model receiving only link $l_i$ attributes cannot distinguish between them. By processing the full ordered sequence of links from origin to destination, DEP resolves this ambiguity and reduces the associated prediction uncertainty.}

\begin{figure}[!h]
  \centering
  \begin{minipage}{0.33\textwidth}
        \centering
        \includegraphics[width=1\textwidth]{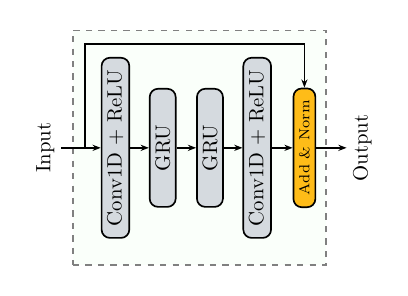}
        \caption{GRU block}
        \label{fig:gruBlock}
  \end{minipage}\hfill
  \begin{minipage}{0.33\textwidth}
        \centering
        \includegraphics[width=1\textwidth]{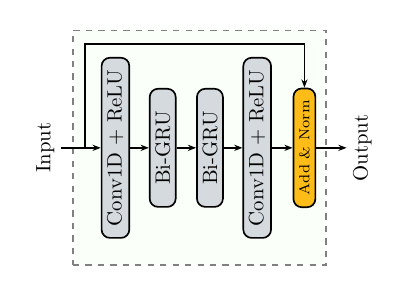}
        \caption{Bi-GRU block}
        \label{fig:biGruBlock}
  \end{minipage}
  \begin{minipage}{0.33\textwidth}
        \centering
        \includegraphics[width=1\textwidth]{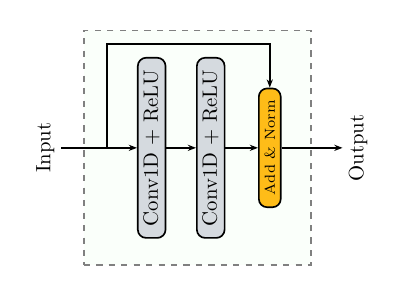}
        \caption{Conv1D block}
        \label{fig:convBlock}
  \end{minipage}
\end{figure}

To further reduce the number of parameters, the model employs GRU and bidirectional Gated Recurrent Unit (Bi-GRU) blocks, presented in \autoref{fig:gruBlock} and \autoref{fig:biGruBlock}, respectively. These blocks utilize a one-dimensional convolutional layer (Conv1D) with a Rectified Linear Unit (ReLU) activation function as the initial step for dimensionality reduction. Following this, GRU and Bi-GRU layers are applied to capture temporal dependencies within the reduced-dimensional space. Subsequently, another Conv1D layer restores the dimensionality to its original size, facilitating the integration of residual connections \citep{resnet}, which require consistent dimensionality across layers outputs. Residual connections are combined with layer normalization \citep{layerNorm} to accelerate and stabilize the training and improve overall model performance. \autoref{fig:convBlock} illustrates the convolutional block employed in the model, which consists of two Conv1D layers, each with a ReLU activation function. Additionally, the block incorporates a residual connection and layer normalization. All Conv1D layers in the model have a kernel size of one, functioning as point-wise fully connected layers.

The model architecture, as depicted in \autoref{fig:model-architecture}, accommodates three distinct types of inputs, each corresponding to different data categories:
\begin{itemize}
    \item $Input_1$: This is the sequential input representing the trajectory of a vehicle during a trip. It includes an ordered sequence of links traversed by the vehicle, along with associated attributes (link length, average speed, square of the average speed, road slope, and speed limit).
    \item $Input_2$: This input comprises vehicle specifications, including weight, engine power, Euro norm, vehicle category, and fuel type, as well as ambient temperature. All categorical variables are encoded using one-hot encoding.
    \item $Input_3$: This input captures temperature conditions, specifically ambient temperature, initial engine temperature, and initial temperature of the exhaust aftertreatment devices.
\end{itemize}  

$Input_1$ is first passed through a Conv1D layer, followed by a Bi-GRU block to capture temporal dependencies across links, producing an output of shape $(n_s, 64)$, where $n_s$ is the number of links. In parallel, $Input_2$ is processed through a two-layer MLP with ReLU activation, then repeated to match the sequence length, also yielding an output of shape $(n_s, 64)$.

These outputs are concatenated into a single sequence of shape $(n_s, 128)$, which is further processed by a Bi-GRU followed by two GRU layers to capture interactions between spatial and contextual features. The GRUs are initialized using $Input_2$ and the last input $Input_3$, processed via a two-layer MLP.

Final emissions predictions at each link are made using two convolution blocks and a final convolution layer. Although ReLU was initially considered for the output layer due to the non-negative nature of emissions, it led to underestimation, especially for NOx,  as low NOx values were frequently predicted as zeros. This phenomenon has also been observed in \citet{good_metrics}. To address this, a linear activation is used, ensuring non-negative and accurate predictions for CO\textsubscript{2} and NOx emissions per link.

For a concise mathematical representation, the model’s predictive function $f$ estimates the CO\textsubscript{2} and NOx emissions, denoted as $y_i = (\text{CO}_{2,i}, \text{NOx}_i)$, for a given road link $i$ as follows:
\begin{equation}
y_i = f(X_{1,0:n}, X_2, X_3), \quad i \in {0, \ldots, n},
\end{equation}
where $X_1$, $X_2$, and $X_3$ correspond to the inputs ($Input_1$, $Input_2$, $Input_3$), with $X_{1,0:n}$ representing the sequence of link-level features for all $n$ links of a trip.

This model has been updated from the version described in \cite{emgen}. The key changes include replacing the LSTM units with GRU blocks to improve computational efficiency. Additionally, vehicle features ($Input_2$) are now integrated directly into the sequence, rather than just being used to initialize the initial states as in the previous version. These updates have led to better accuracy for very long trips and improved the accuracy of cold start emissions modeling. Moreover, initial temperature of the exhaust gas aftertreatment technology and ambient temperature have been added to the inputs due to their impact on emissions. The dataset has also been updated following further calibrations on MOBSim.

{
Compared to traditional mesoscopic and macroscopic models such as COPERT and HBEFA, which require only average speed and vehicle category, DEP incorporates a richer set of inputs including vehicle weight, engine power, road slope, speed limit, and initial temperature conditions. While these inputs are not universally available in all deployment contexts, they are increasingly accessible through national vehicle registration databases, and commercial road network providers (e.g., HERE, OpenStreetMap). The MATSim integration described in this study represents one instantiation of DEP's workflow, but the model is not inherently dependent on MATSim and can be applied with any data source that provides link-level average speeds and the associated attributes. The main idea behind this model is to be able to substitute a microscopic model (SPG-M) with a more efficient model without and lose of granularity or accuracy. In this case, when compared to microscopic emission models, we are not using inputs that are not used in the microscopic emission model used in SPG-M, this is MOBSim.
}

\subsubsection{Training Procedure}
The prepared dataset is randomly split into training, validation, and test datasets. The proportions of each set are 90\%, 5\%, and 5\% respectively. The training dataset is used for the training. The validation dataset is used to evaluate the developed models and find the optimal hyperparameters, such as the number of layers, number of units per layer, or modifications in the architecture. The test dataset is employed to assess the model's performance after training to ensure it generalizes well to new data.

\paragraph{Loss Function}
The Mean Squared Error (MSE) is selected as the loss function for this study. MSE is a standard choice for regression problems in deep neural network frameworks due to its effectiveness in emphasizing larger errors and its computational efficiency.
\paragraph{Optimizer}
The Adam optimizer \citep{adamOptimizer} is employed for its adaptability in handling sparse gradients on noisy problems. To optimize convergence, a dynamic learning rate adjustment strategy was applied, specifically halving the learning rate every 10 epochs throughout a total of 100 epochs.
\paragraph{Regularization Techniques}
To mitigate the risk of overfitting and enhance the generalization ability of the model, a dual regularization strategy using both L1 and L2 norms is implemented \citep{ying2019overview}. Additionally, Gaussian noise layers with zero mean were integrated into the network architecture. These noise layers serve as a form of stochastic regularization, improving the model's robustness.
\paragraph{Weight Pruning}
To further increase computational efficiency and reduce model complexity, a threshold-based weight pruning method was adopted. Weights (model parameters) with an absolute value below $10^{-2}$ were pruned, eliminating negligible parameters that contribute minimally to the predictive capabilities of the model. Post-training, the model comprises $210 \thinspace 172$ parameters, of which $44 \thinspace 454$ ($21.15\%$) are zeros,

\subsubsection{Results}
This section provides a detailed evaluation of the model's performance in estimating CO\textsubscript{2}  and NOx emissions, structured into two parts. First, the overall accuracy of the model is assessed in estimating emissions. Second, a comparative analysis is performed against PEMS data, serving as a benchmark to validate the model's predictions. 

\paragraph{Model Evaluation}

The performance of the model is assessed using a suite of evaluation metrics. First, the Total Percentage Error (TPE) is employed, which is commonly used in the literature to assess the predictive accuracy of NOx emissions models, as discussed in \citet{good_metrics}. TPE quantifies the overall discrepancy between the observed and predicted values across the entire dataset, thus representing the proportion of emissions captured by the model. It is calculated using the following formula:

\begin{equation}
    TPE(\%) = \frac{\sum Y_{predicted} - \sum Y_{actual}}{\sum Y_{actual}} \times 100 .
\end{equation}

In this formula, $\sum Y_{predicted}$ represents the total predicted emissions across the dataset, while $\sum Y_{actual}$ represents the total actual emissions.

However, the TPE should not be used as the only metric. That's why the coefficient of determination, denoted by $R^2$, is also used to assess the model's predictive accuracy. $R^2$ measures the proportion of variance in the dependent variable explained by the independent variables. A higher $R^2$ value indicates a stronger correlation between predicted and actual emissions, reflecting better model performance.

To further analyze the model's effectiveness in capturing the underlying patterns that drive emissions, Spearman's Rank Correlation Coefficient ($r_s$) is employed. This non-parametric metric assesses the strength and direction of the monotonic relationship between variables, offering insight into the model's ability to replicate the relative ordering of emissions.

These metrics are applied across the three datasets, including training, validation, and test sets. {It should be noted that these datasets are MOBSim-generated targets from the synthetic training dataset, not real-world measurements.} The evaluation is conducted at both the link level and trip level, allowing for a detailed examination of the model's capacity to represent the spatial distribution of emissions throughout a trip.  For the link-level analysis, emissions of NOx are quantified in milligrams (mg), and CO\textsubscript{2} in grams (g). At the trip level, the emissions are calculated as milligrams per kilometer (mg/km) for NOx and grams per kilometer (g/km) for CO\textsubscript{2}. These units are used in the computation of the statistical metrics.


The results are presented in \autoref{tab:metrics}, which demonstrates the model's performance in estimating emissions across the three datasets. The model performs particularly well with unseen data, as indicated by the closely matching metrics across all datasets. It consistently achieves high scores for both \(R^2\) and \(r_s\), highlighting its ability to estimate both NOx and CO\textsubscript{2} emissions at the link and trip levels. The slight outperformance on the test dataset is due to its higher proportion of Diesel vehicles compared to the other datasets. {DEP performs better for diesel vehicles and for Euro norms up to 5 than for Euro 6d vehicles, whose NOx emissions are significantly lower and more sensitive to driving dynamics and temperature. Small variations in fleet composition across splits therefore cause slight metric differences.}

\begin{table}[h]
\centering
\caption{Evaluation metrics for train, validation and test set}
\label{tab:metrics}
\small
\begin{tabular}{ c c c c c }
\Xhline{1.5pt}
\multirow{2}{*}{Prediction Type} & \multicolumn{2}{c}{Link Level} & \multicolumn{2}{c}{Trip Level} \\
\cline{2-5}\\
& CO\textsubscript{2}  & NOx & CO\textsubscript{2}  & NOx \\
\hline
Train TPE (\%)  & -0.75 & -1.72 & -0.87 & -1.90 \\
Val TPE (\%) & -0.82 & -1.74 & -0.83 & -1.89\\
Test TPE (\%)  & 0.10 & -0.49 & 0.02 & -0.52 \\
\hline
Train $R^2$ & 0.942 & 0.935 & 0.928 & 0.987 \\
Val $R^2$  & 0.943 & 0.937 & 0.930 & 0.988 \\
Test $R^2$ & 0.942 & 0.924 & 0.911 & 0.983 \\
\hline
Train $r_s$ & 0.960 & 0.919 & 0.955 & 0.993 \\
Val $r_s$  & 0.957 & 0.919 & 0.955 & 0.994\\
Test $r_s$ & 0.976 & 0.920 & 0.945 & 0.993 \\
\Xhline{1.5pt}
\end{tabular}
\end{table}

\begin{figure}[h]
    \centering
    \includegraphics[width=0.7\textwidth]{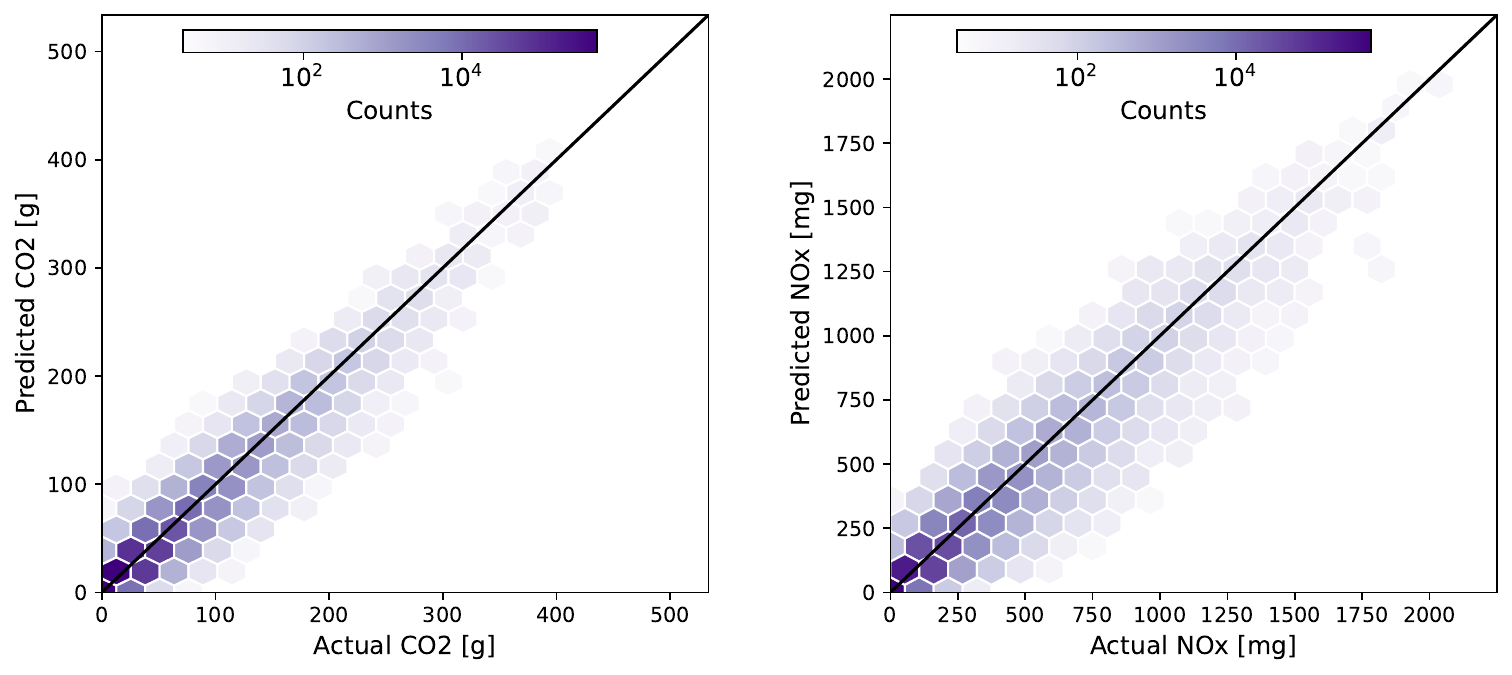}
    \caption{Hexbin plot of actual vs. predicted emissions (link level)}
    \label{fig:linklevelco2}
\end{figure}

\begin{figure}[h]
    \centering
    \includegraphics[width=0.7\textwidth]{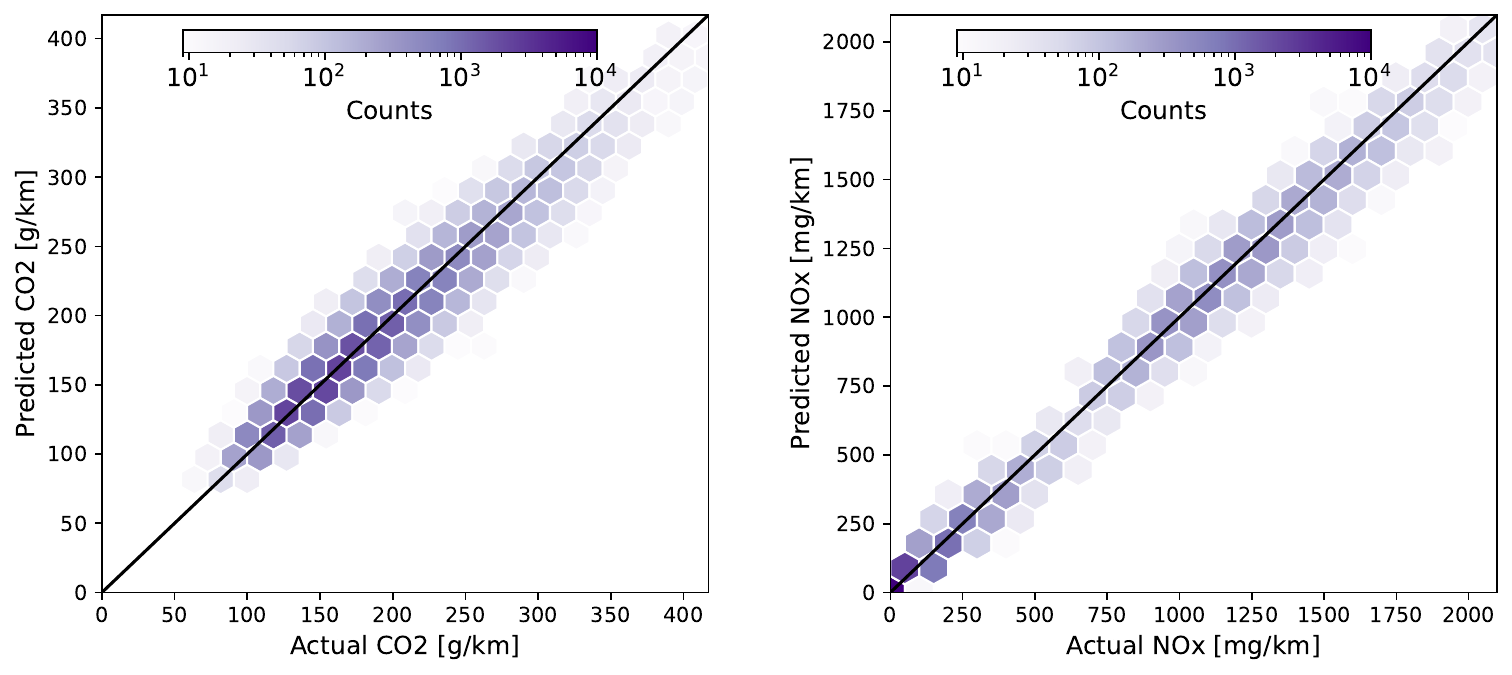}
    \caption{Hexbin plot of actual vs. predicted emissions (trip level)}
    \label{fig:triplevelCO2}
\end{figure}

{A hexbin density plot comparing actual and predicted CO\textsubscript{2} and NOx values at the link level using the test dataset is shown in \autoref{fig:linklevelco2}. The plot indicates that most frequently occurring prediction-reference pairs lie near the y = x line, demonstrating a strong correlation between reference and predicted values. At the trip level, \autoref{fig:triplevelCO2} shows slightly lower dispersion in NOx predictions, as positive and negative errors at the link level can, in some cases, cancel each other out at the trip level.}

\begin{figure}[h]
    \centering
    \includegraphics[width=0.8\textwidth]{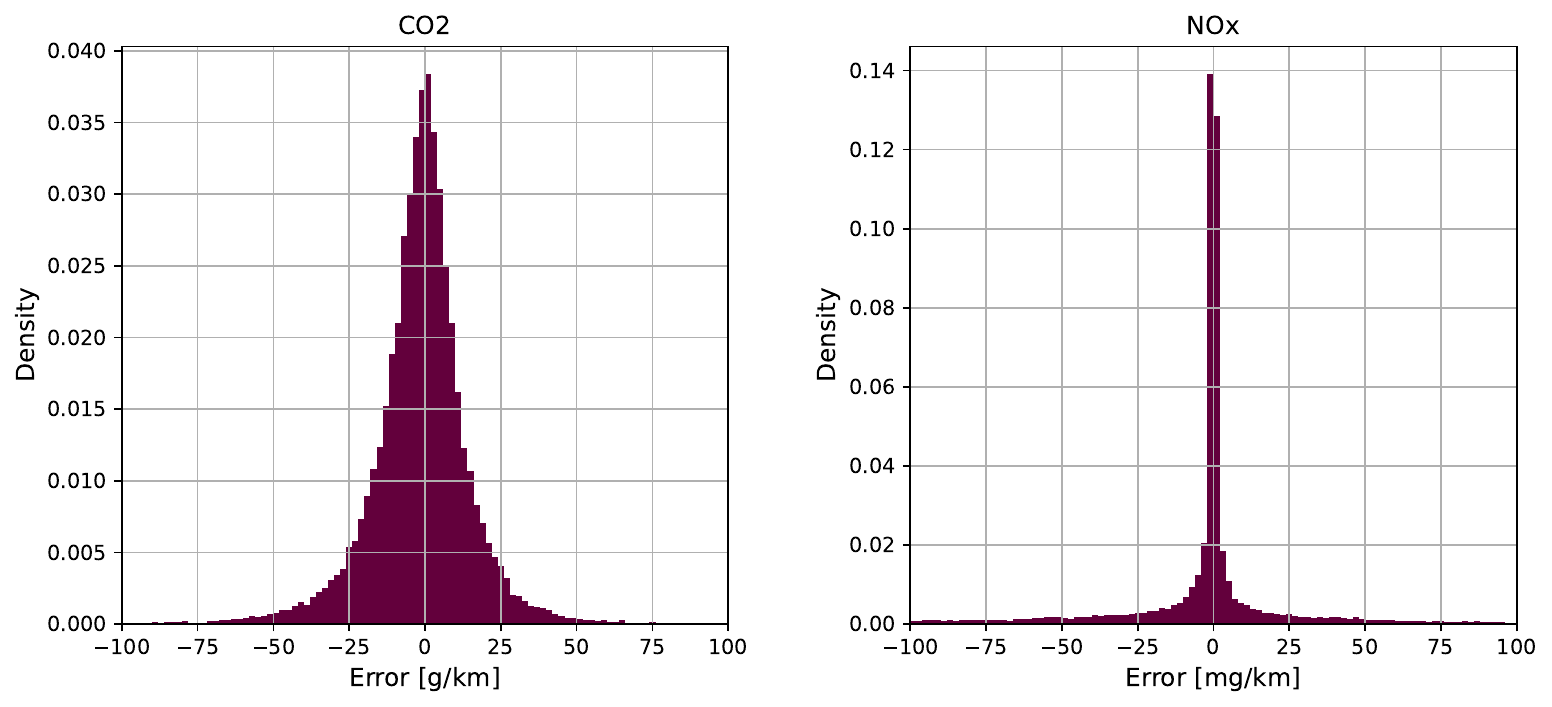}
    \caption{CO\textsubscript{2} and NOx error distribution at the trip level}
    \label{fig:tripLevelErrors}
\end{figure}

The distribution of errors differs between CO\textsubscript{2} and NOx emissions, as illustrated in \autoref{fig:tripLevelErrors}. The error distribution for NOx exhibits a high density near zero, whereas the error distribution for CO\textsubscript{2} approximates a normal distribution. This discrepancy arises from the inherent characteristics of the data. While vehicles tend to have comparable CO\textsubscript{2} emission levels, their NOx emission levels vary significantly. As a result, the NOx error distribution shows a high density near zero, particularly because the data includes a mix of both high-emission and low-emission vehicles, with the latter skewing the errors toward zero.

\paragraph{Model Predictions Compared to PEMS measurements}

The primary objective of this section is to further validate the model by evaluating its predictive performance against experimental emission data. It is important to note that, since the model is trained on data generated by MOBSim, its ability to predict emissions is inherently constrained by the accuracy of MOBSim itself. Consequently, a portion of the prediction error originates from limitations in MOBSim, and the model's accuracy cannot exceed that of the simulation framework on which it is based.

Experimental datasets from various projects conducted by IFPEN are used for this evaluation \citep{mobsim2, gecoAiremissions, mobsim4}. To provide a benchmark, the model's performance is compared with that of COPERT v5.7.

The evaluation process involved selecting a set of trips that included both test bench and on-road measurements for diesel and gasoline vehicles. The available dataset contains speed, slope, emissions, and vehicle characteristics. However, DEP requires input data aggregated at the link level. This presents a challenge, as the dataset lacks predefined links and cannot be segmented manually due to the absence of link-related information.

The model is specifically trained to predict emissions at the link level, where emissions are strongly influenced by how these links are defined. Link definitions are based on features such as intersections, speed limit changes, or the presence of traffic lights, elements that shape speed patterns and, consequently, affect emission levels, as shown in \citet{moh}. Therefore, the model’s performance is highly dependent on accurate link definitions and their associated attributes.

In addition, the dataset does not provide speed limit information, which is a required input for the model. To overcome this, the trips were divided into equidistant links of 80 meters, and slope and speed were aggregated within each link. Speed limits were approximated based on the average speed of each link.

It is important to keep in mind that the simplified, equidistant link segmentation used here does not fully capture the complexity of the original link definitions, potentially affecting the accuracy of the model’s predictions. Despite these limitations, the analysis offers useful insights into the model’s ability to estimate emissions under real-world conditions. 

For comparison, both MOBSim and COPERT are employed: MOBSim uses the measured second-by-second speed profiles and vehicle characteristics as input, while COPERT relies on the average speed and vehicle attributes.

\begin{figure}[h]
    \centering
    \vspace{-0.1cm}
    \begin{tikzpicture}
        \node[anchor=south west, inner sep=0] (image) at (0,0) {
            \includegraphics[width=0.8\textwidth]{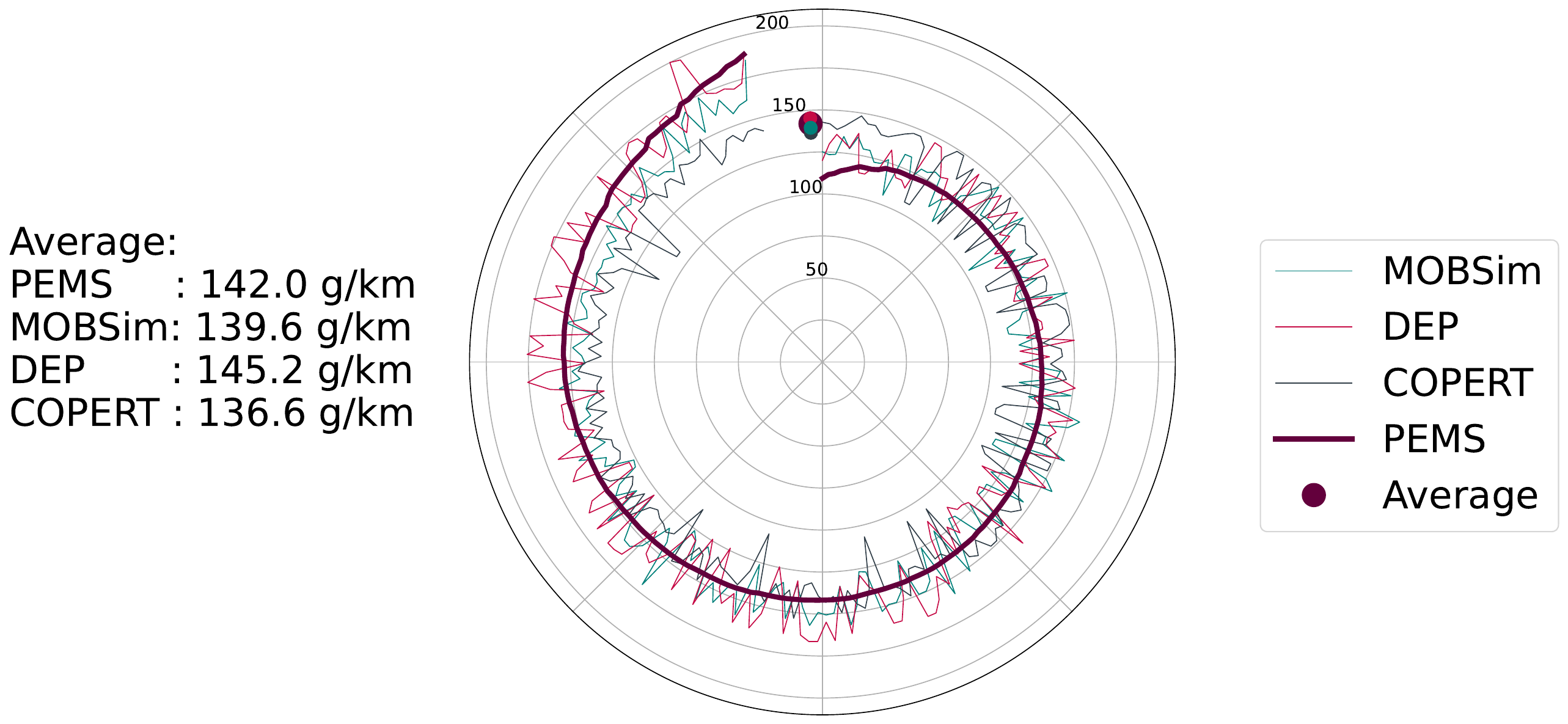}
        };        
        \begin{scope}[x={(image.south east)}, y={(image.north west)}]            
            \draw[thick,->,black] (0.525,1) -- (0.525,1.12); 
            \node[rotate=90,anchor=south] at (0.525,1.06) {\scriptsize g/km};
            
            \draw[thick,->,black] (0.53,1.01) to[out=0, in=140]  (0.72,0.81);
            \node[rotate=-37,anchor=west] at (0.665,0.95) {\scriptsize trips};
        \end{scope}
    \end{tikzpicture}
    \caption{Comparison of Model CO\textsubscript{2} Predictions (g/km) with COPERT, MOBSim, and PEMS Measurements}
    \label{fig:withPEMS}
\end{figure}

\autoref{fig:withPEMS} presents the results of CO\textsubscript{2}  emissions across 192 trips, having an average distance of 74 km. Of these, 105 trips were conducted in gasoline vehicles and 89 trips in diesel vehicles. Although the predictions exhibit oscillations around the measured values, a key observation from this figure is that both MOBSim and DEP capture the overall trend of the measured emissions, unlike COPERT, which oscillates within a narrow range and fails to follow the trend observed in the measurements. {The Root Mean Square Error (RMSE) is 12.8 g/km for MOBSim, 14.0 g/km for DEP, and 20.1 g/km for COPERT}. This indicates the superior capability of DEP in predicting CO\textsubscript{2} emissions compared to COPERT, despite the difficulties previously discussed. {This is particularly evident when trips are ordered by increasing emissions: DEP and MOBSim closely track this trend, whereas COPERT remains within a narrow prediction band regardless of the actual emission level, confirming that average-speed-based models capture aggregate totals but not local variability.}

It is important to note, however, that emissions are mostly overestimated by DEP compared to MOBSim, a pattern also observed in the average CO\textsubscript{2} emissions across all trips. This overestimation was not present in any of the test or validation datasets and may be attributed to the links definition. Additionally, part of the dispersion in errors from both MOBSim and DEP can be partially attributed to the absence of certain input data, such as temperature values in some test cases.

A noteworthy observation is that COPERT demonstrates acceptable predictive performance when evaluating average emission levels across all measurements, despite exhibiting a slight tendency to underestimate. This is largely due to the long distances covered by these trips and the diversity of vehicles and driving conditions, including both urban and extra-urban environments. COPERT's ability to capture average CO\textsubscript{2} emissions over long distances has been previously recognized in the literature \citep{samaras2019improvedCopert}. However, it is crucial to emphasize that COPERT's performance diminishes when applied at the link level, particularly in scenarios characterized by shorter segments, which are common in urban networks where link-level emissions estimation is required. {It should be noted, however, that the vehicle sample used in this comparison consists exclusively of Euro 6 compliant vehicles, which may not be representative of the broader EU fleet that COPERT is calibrated to model. A more diverse sample, including older Euro norms and a wider range of driving conditions, might yield different relative performance across models. The intent of this comparison is therefore not to draw general conclusions about COPERT's accuracy, but to illustrate the behavior of each model on the available experimental data and to highlight the structural limitation of average-speed-based approaches in capturing emission variability across individual trips and links.}

\begin{figure}[h]
    \centering    
    \includegraphics[]{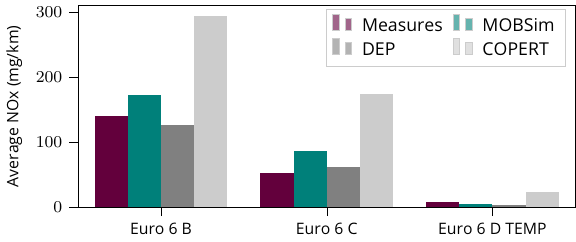}
    \caption{Comparison of model predictions for NOx emissions (mg/km) with COPERT, MOBSim, and PEMS measurements. Diesel vehicle share within each Euro norm: Euro 6B: 65\% diesel, Euro 6C: 36\% diesel, Euro 6D-temp: 0\% diesel (all gasoline).}
    \label{fig:NOxMeasures}
\end{figure}

Evaluating the model's performance in estimating NOx emissions is more complex due to the significant variability in NOx emissions among vehicles within the same category and compliant with the same Euro norm. For instance, as highlighted in \citet{petroviceuropean}, NOx emissions can range from 450 to over 1800 mg/km within Euro 5 vehicles. Given this variability, it is more insightful to evaluate the average NOx emissions rather than individual values to determine if the model can reproduce consistent emission levels. The results are illustrated in \autoref{fig:NOxMeasures}.

It is important to note that all the measurements in this study were conducted on Euro 6 compliant vehicles, which are less polluting compared to older models but have NOx emissions that are more sensitive to factors such as acceleration and temperature. A notable characteristic of COPERT when applied to Euro 6 vehicles is that, despite its general tendency to underestimate NOx emissions according to several studies \citep{underEstimationNOx1, underEstimationNOx3}, it actually overestimates NOx emissions for these vehicles, as demonstrated in \citet{theseEmissionsLondon} and further supported by the results of this study. 

The results indicate that MOBSim overestimates NOx emissions for these simulated vehicles, although it performs better than COPERT. {However, this should not be interpreted as a general tendency for MOBSim to overestimate NOx emissions. Instead, it underestimates NOx emissions specifically for the Euro 6 vehicles for which PEMS measurements are available in this study; whether this tendency extends to other Euro norms cannot be assessed from the current dataset}. DEP appears to outperform MOBSim, but a more nuanced interpretation is that DEP underestimates NOx emissions relative to MOBSim (because it was trained on datasets generated using MOBSim), which, by coincidence, is closer to the measured values. This discrepancy between MOBSim and DEP can be explained similarly to the discrepancies observed with CO\textsubscript{2} emissions.

In conclusion, the results indicate that while both MOBSim and DEP outperform COPERT in estimating NOx emissions, there remains considerable room for improvement. Increasing the accuracy of NOx emission modeling may be achieved by integrating a larger dataset of on-road NOx measurements and incorporating additional relevant parameters currently not accounted for.

It is worth noting that the applicability of this emission model across different regions may require the creation of a new emission dataset and retraining of the model. Emission models can be highly region-specific due to differences in vehicle fleets, driving conditions, and emission regulations. Applying a model developed for one region to another without appropriate adjustments may lead to significant inaccuracies \citep{theseEmissionsLondon, liu2013vehicle}. {In addition, in contexts where detailed vehicle attributes are unavailable, we recommend using a representative vehicle rather than population-average input values. This is because the emission response to weight and power is nonlinear: averaging inputs does not yield the same emission prediction as a vehicle whose emissions are representative of the fleet average for that category. A representative vehicle can be defined as the vehicle within a given fuel-type and Euro-norm segment whose predicted emissions most closely match the segment's average emission level.}

\subsection{Workflow Integration}
The workflow illustrated in \autoref{fig:workflow} is integrated as follows. After the MATSim simulation, all car trips are collected. For each trip, the vehicle is tracked from its origin to its destination, and its average speed on each road link is recorded together with the corresponding link attributes.

Although applying the DEP model may appear straightforward once all trips are available, a key challenge arises because DEP is trained on real-world speed data that reflect realistic speed dynamics between links—dynamics that are not captured by MATSim and must be explicitly corrected.

Moreover, while DEP enables direct emission estimation, accurately accounting for cold-start emissions requires knowledge of the engine’s initial temperature. Ambient temperature can serve as a rough approximation, but in real-world conditions, vehicles often begin trips with engines already warmed up from prior use. Therefore, an additional model is needed to estimate the engine’s initial temperature at the start of each trip.

\subsubsection{Speed Filter}
Since the proposed emission model accounts for temporal dependencies within trips, an accurate representation of speed, particularly average speed variations between consecutive links, is essential. Large fluctuations in speed between links result in significant accelerations and decelerations, which are associated with higher emissions. Conversely, smoother transitions generally lead to lower emissions.

Our previous work has shown that this constitutes a limitation of MATSim \citep{spgm}. For instance, under free-flow conditions, the MATSim traffic model assumes vehicles travel at the free-flow speed of each link without considering the speed carried over from the previous link. This simplification can lead to unrealistically high accelerations when transitioning between low-speed and high-speed links.

Moreover, it has been demonstrated in \citet{triebke2023bridging, triebke2020pre} that MATSim's queueing model can produce unrealistic congestion patterns under certain conditions, particularly on short links (common in urban environments) or when using small sample sizes. A third contributing factor is numerical congestion caused by the discrete time-stepping nature of MATSim, typically operating at 1-second intervals.

To address these limitations, a post-processing approach was previously proposed: average speeds per link across an entire trip (from origin to destination) are first corrected, then smoothed through a filtering technique. In this study, we apply the same speed correction and filtering methodology that is applied in \citet{spgm}. For reference, the filter is defined as follows:

\begin{equation}\label{equaFilter}
    a = \max \left( a_{\text{max}} \left[ 1 - \left( \frac{v}{\mathcal{M}_w(v_{0})} \right)^\delta \right] , a_{\text{min}} \right),
\end{equation}

where $\mathcal{M}_w$ denotes a moving average filter with a window size of $w$ meters. The parameters $a_{\text{max}}$ and $a_{\text{min}}$ represent the maximum allowable acceleration and minimum allowable deceleration, respectively. $v_0$ is the original speed from MATSim, $v$ is the filtered speed, and $a$ is the filtered acceleration, defined as the time derivative of filtered speed. The application of the filter and the calibration procedure are detailed in \citet{spgm}. The the final calibrated values are $ (a_{\text{max}}, a_{\text{min}}, w) = (-2.3, 2.3, 50) $.

\subsubsection{Initial Temperatures}
Knowledge of the initial engine temperature is required to estimate cold-start emissions with DEP accurately. A modified version of Newton’s Cooling Law (NCL) proposed in \citet{MNCL}, and calibrated in \citet{spgm}, is used for this purpose. The model is given by:

\begin{equation} \label{equa:initialEngineTemperature}
     T(t) = T_a + (T_0 - T_a) e^{\lambda t^{\psi}},
\end{equation}

where $ T(t) $ denotes the engine temperature after a period $t$ following engine shutdown, $ T_a $ is the ambient temperature, $ T_0 $ is the engine temperature at shutdown, and $ \lambda $ and $ \psi $ are model parameters.

Calibration details and validation against experimental data are provided in \citet{spgm}. The calibrated parameter values are $ (\lambda, \psi) = (-1.83, 0.79) $. In the workflow, this model is used to estimate the initial engine temperature following the MATSim simulation, based on the ambient temperature, the engine temperature at the end of the previous trip (assumed to be 90 °C, representing a fully warmed-up engine), and the duration between the end of the previous trip and the start of the current trip. For the first trip of the day, the ambient temperature is used as the initial engine temperature.

{The same cooling law (Equation \ref{equa:initialEngineTemperature}) is applied to estimate the initial temperature of exhaust aftertreatment devices, using a fully warmed-up reference temperature of 300°C instead of the 90°C for the engine. We acknowledge this as a limitation: in reality, aftertreatment device temperatures follow different cooling dynamics than the engine, and a single shared cooling law with a shifted reference temperature is a simplification \citep{ref1reviewer}. However, given the absence of large-scale experimental data on aftertreatment cooling rates, this approximation represents a pragmatic solution.}

\section{Results and Discussion}\label{sec:results}

This section presents emission estimates obtained using the proposed DEP workflow and compares them with those from existing models: SPG-M, COPERT, and HBEFA.

In this study, COPERT v5.7 Tier 3 is implemented to estimate emissions based on the average vehicle speed per link provided by MATSim and vehicle-specific characteristics. HBEFA v4.1, which considers traffic conditions instead of the average speed, is used. Its integration with MATSim is detailed in \citet{hbefaMatsim}. Neither COPERT nor HBEFA accounts for road gradient effects or detailed cold start emissions. Therefore, their evaluation in this section is limited to hot emission scenarios.

The SPG-M workflow, developed in previous work \citep{spgm}, generates a detailed speed profile for each trip simulated by MATSim, using a deep learning based speed profile generator. Emissions are then estimated using MOBSim. In comparison to the workflow presented in \autoref{fig:workflow}, the only difference lies in the emission estimation component. DEP and SPG-M are expected to produce equivalent results, as they are both based on MATSim and MOBSim.

The analysis includes global emission comparisons, validation against on-road remote sensing data, spatial distribution assessments, and cold start emission analysis. The computational efficiency of the models is also evaluated.

\subsection{Global Emission}

To evaluate the accuracy and efficiency of DEP, it was compared to SPG-M under three operating conditions: fully warmed-up engines (hot), cold starts at 20°C ambient temperature (cold), and hot starts including road gradient effects (slope).

\begin{figure}[h]
    \centering
    \includegraphics[]{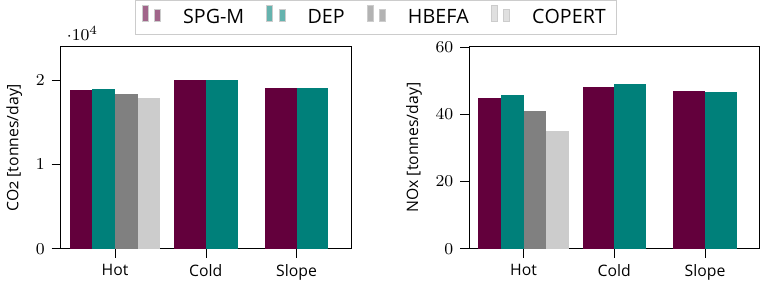}
    \caption{Comparative analysis of CO\textsubscript{2} and NOx emissions estimated by SPG-M and DEP across different scenarios: Hot starts, cold starts, and slope considerations.}
    \label{fig:comparEmgenMobsim}
\end{figure}

As shown in \autoref{fig:comparEmgenMobsim}, CO\textsubscript{2} and NOx emissions estimated by DEP closely match those from SPG-M. Specifically, differences in CO\textsubscript{2} emissions are minimal: 0.27\%, 0.35\%, and -0.06\% for hot, cold, and slope scenarios, respectively. Differences in NOx emissions are slightly larger but remain consistent: 1.67\%, 1.87\%, and -0.27\%.

In contrast, both COPERT and HBEFA underestimate emissions compared to DEP and SPG-M, particularly for NOx. HBEFA predicts NOx emissions approximately 17\% higher than COPERT, a discrepancy also reported in previous studies \citep{borge2012comparisonMadrid}. This difference likely arises because COPERT relies solely on average speed, which does not adequately represent congestion \citep{samaras2019improvedCopert,smit2008air}.

\subsection{Validation Against Real-World Remote Sensing Data}

The relatively higher NOx emissions predicted by DEP and SPG-M compared to COPERT and HBEFA raise questions about their accuracy, especially in the absence of direct ground truth measurements. To evaluate this, we compared SPG-M and DEP estimates with real-world NOx emissions data from remote sensing campaigns conducted by the True Initiative \citep{trueInitiative}.

\begin{figure}[h]
    \centering  
    \includegraphics[]{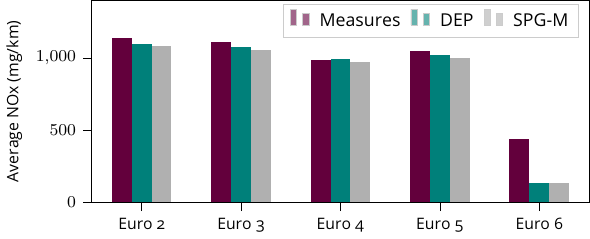}
    \caption{Average NOx emissions of the on-road fleet \citep{trueInitiative}, and of SPG-M, from Euro 2 to Euro 6, for diesel passenger vehicles.}
    \label{fig:NOxEmissionsByEuro}
\end{figure}

{\autoref{fig:NOxEmissionsByEuro} presents average NOx emissions by Euro norm for diesel passenger vehicles. For Euro 2 through Euro 5, DEP and SPG-M show close agreement with remote sensing measurements, with modest underestimation: DEP underestimates by 3.5\%, 2.7\%, 0.2\%, and 2.9\% for Euro 2, Euro 3, Euro 4, and Euro 5, respectively, while SPG-M underestimates by 4.6\%, 4.8\%, 2.0\%, and 4.4\%. For Euro 6 diesel vehicles, however, both models exhibit substantially larger underestimation: DEP predicts 136 mg/km against a measured value of 440 mg/km ($-69.1\%$), and SPG-M predicts 134 mg/km ($-69.5\%$). This pronounced gap is attributable to the composition of the remote sensing dataset, which was collected between 2011 and 2017 and therefore predominantly captures earlier Euro 6b and Euro 6c vehicles, while excluding newer Euro 6d and Euro 6d-temp vehicles, which comply with stricter regulations and emit significantly lower NOx.}

Overall, despite DEP and SPG-M predicting higher emissions than COPERT and HBEFA, their estimates remain below measured values within an acceptable range. This suggests that DEP and SPG-M provide a more realistic approximation of real-world emissions compared to macroscopic emission models.

\subsection{Spatial Distribution of Emissions}

To evaluate the local-scale performance of DEP and its ability to reproduce local emission peaks, while also highlighting why COPERT cannot be used to analyze such peaks, we first examine the emissions from one of the simulated trips, from its origin to its destination. This allows us to assess how low- and high-emission links are captured by DEP in comparison to SPG-M, and to show how COPERT fails to do so. Second, we analyzed the spatial distribution of daily NOx emissions in the Idf region under the hot scenario and compare it to SPG-M predictions.

\begin{figure}[!b]
    \centering
    \includegraphics[width=0.7\linewidth]{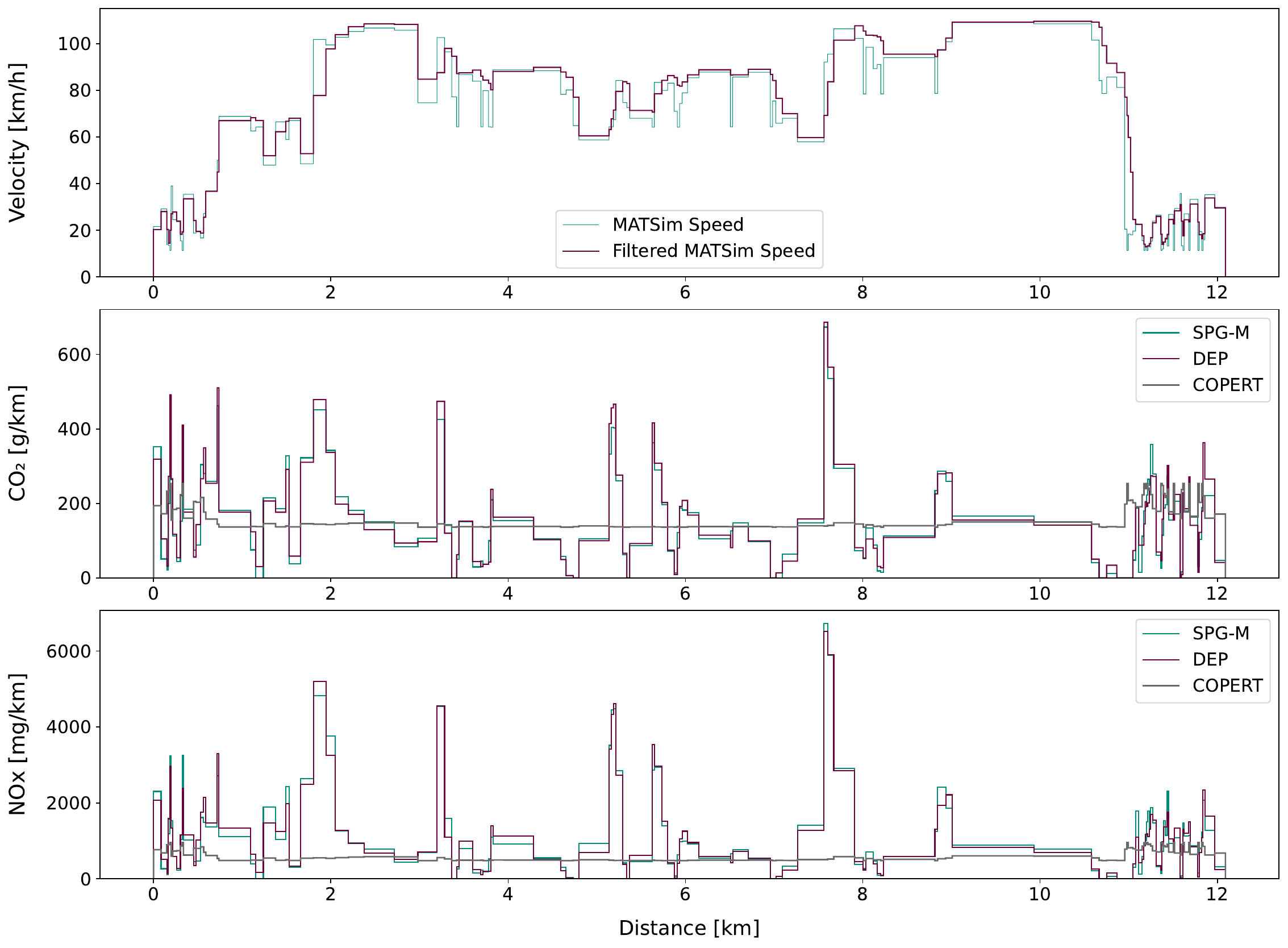}
    \caption{Emission estimates (CO\textsubscript{2}, NOx) of a simulated trip using DEP, SPG-M, and COPERT}
    \label{fig:speedProfile}
\end{figure}

\autoref{fig:speedProfile} illustrates an example trip, showing the average speed per link from origin to destination as simulated by MATSim, along with the corresponding filtered speed. The figure also presents $CO_2$ and NOx emissions estimated at the link level by DEP, SPG-M, and COPERT. This comparison demonstrates the ability of the proposed methodology to capture the spatial variability of emissions more effectively than a macroscopic model. The emissions estimated by COPERT exhibit a relatively flat profile, failing to reflect the real variations that occur along the trip. In contrast, DEP produces emission patterns closely aligned with those of SPG-M, meeting the objectives of this study, and successfully identifies both high- and low-emission zones. These include areas of high acceleration, such as transitions from low-speed to high-speed roads, as well as regions with low emissions, while maintaining realistic overall emission levels.

\begin{figure}[h]
    \centering
    \includegraphics[width=1\textwidth]{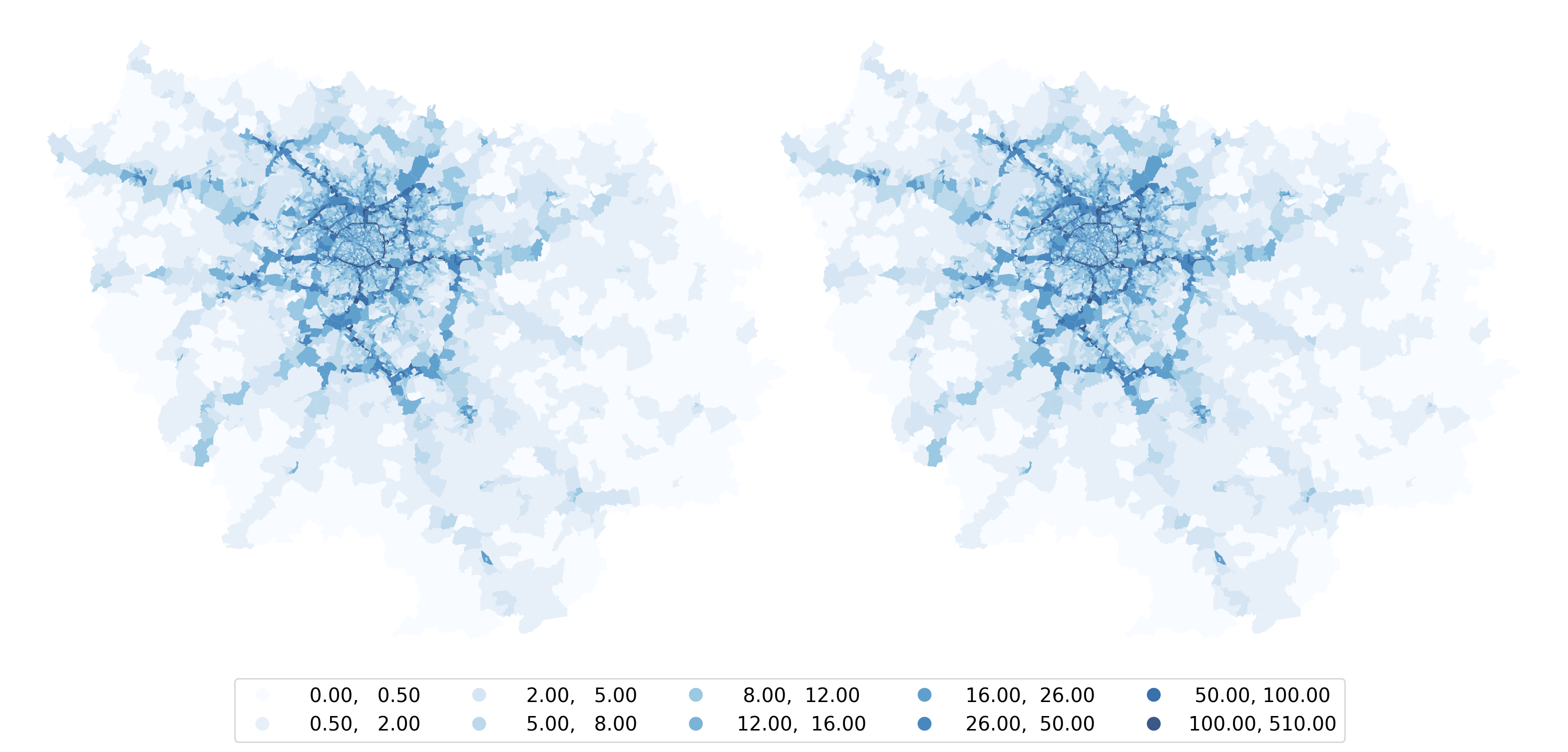}
    \caption{Daily NOx emissions in mg/m² at the IRIS level: Comparison of estimates by SPG-M (left) and DEP (right).}
    \label{fig:comparEmgenMobsimNox}
\end{figure}

\autoref{fig:comparEmgenMobsimNox} presents the spatial distribution of NOx emissions at the IRIS level, a French administrative zoning unit. Emissions estimated by DEP closely match those of SPG-M, indicating strong consistency in local emission estimates.

\begin{figure}[h]
    \centering
    \includegraphics[width=0.8\textwidth]{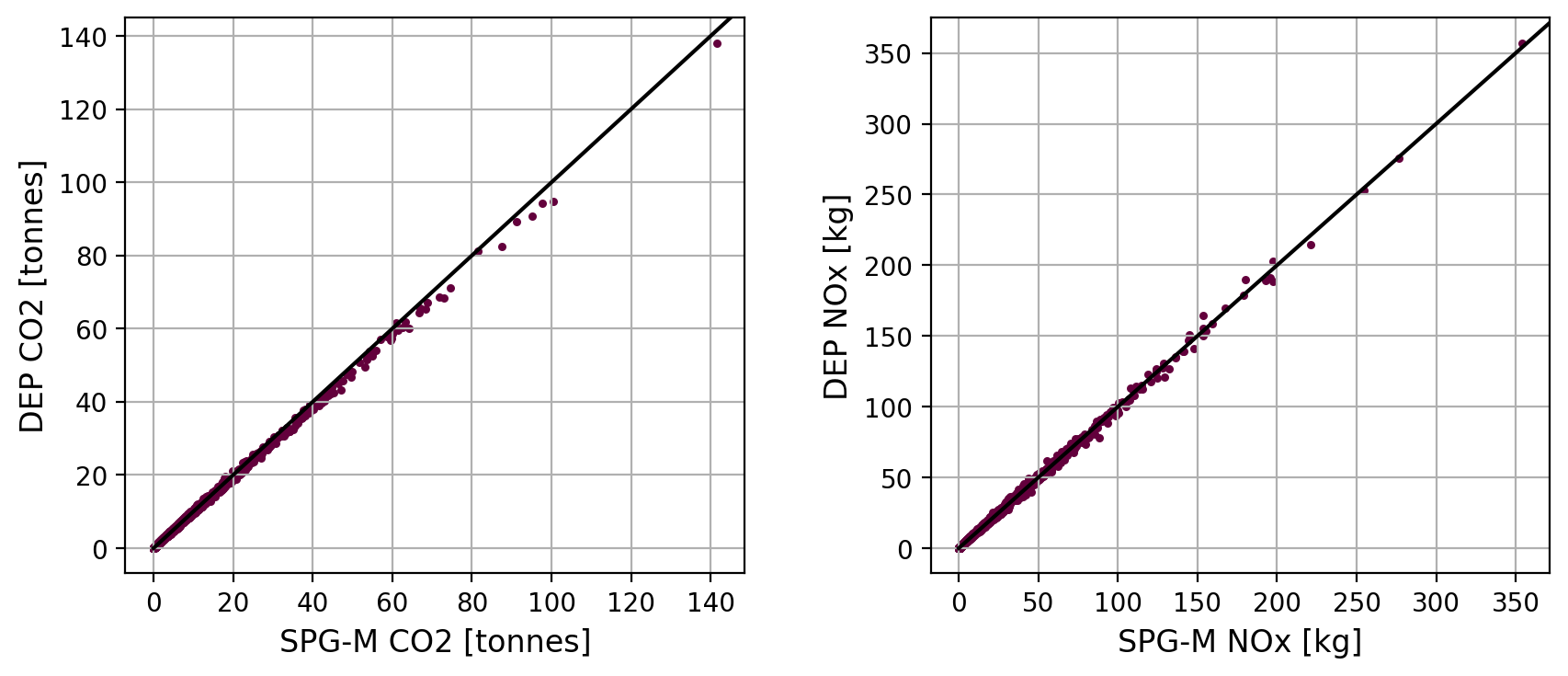}
    \caption{Comparative analysis of daily CO\textsubscript{2} and NOx emissions estimated by SPG-M and DEP at the IRIS level (each point is an IRIS zone).}
    \label{fig:comparEmgenMobsimIris}
\end{figure}

Further validation is provided by \autoref{fig:comparEmgenMobsimIris}, which shows a scatter plot comparing emission levels at the IRIS level. This comparison demonstrates close alignment between the two models, with minimal discrepancies even at this granular scale. These results suggest that DEP is robust and reliable across diverse urban configurations, supporting its use as a viable alternative to SPG-M for both large-scale and localized emission assessments.

\subsection{Spatial Distribution of Cold Start Emissions}

Conventional macroscopic models vary in their treatment of cold-start emissions: some provide a total value, while others do not. However, none specify the spatial or temporal distribution, that is, where or when these emissions occur along a trip. In contrast, DEP explicitly tracks cold-start emissions by following each vehicle throughout its entire trip from origin to destination, enabling detailed analysis of their contribution to total emissions, both globally and locally.

\begin{figure}[h]
    \centering
       \begin{tikzpicture}
        \node[anchor=south west, inner sep=0] (image) at (0,0) {\includegraphics[width=0.95\textwidth]{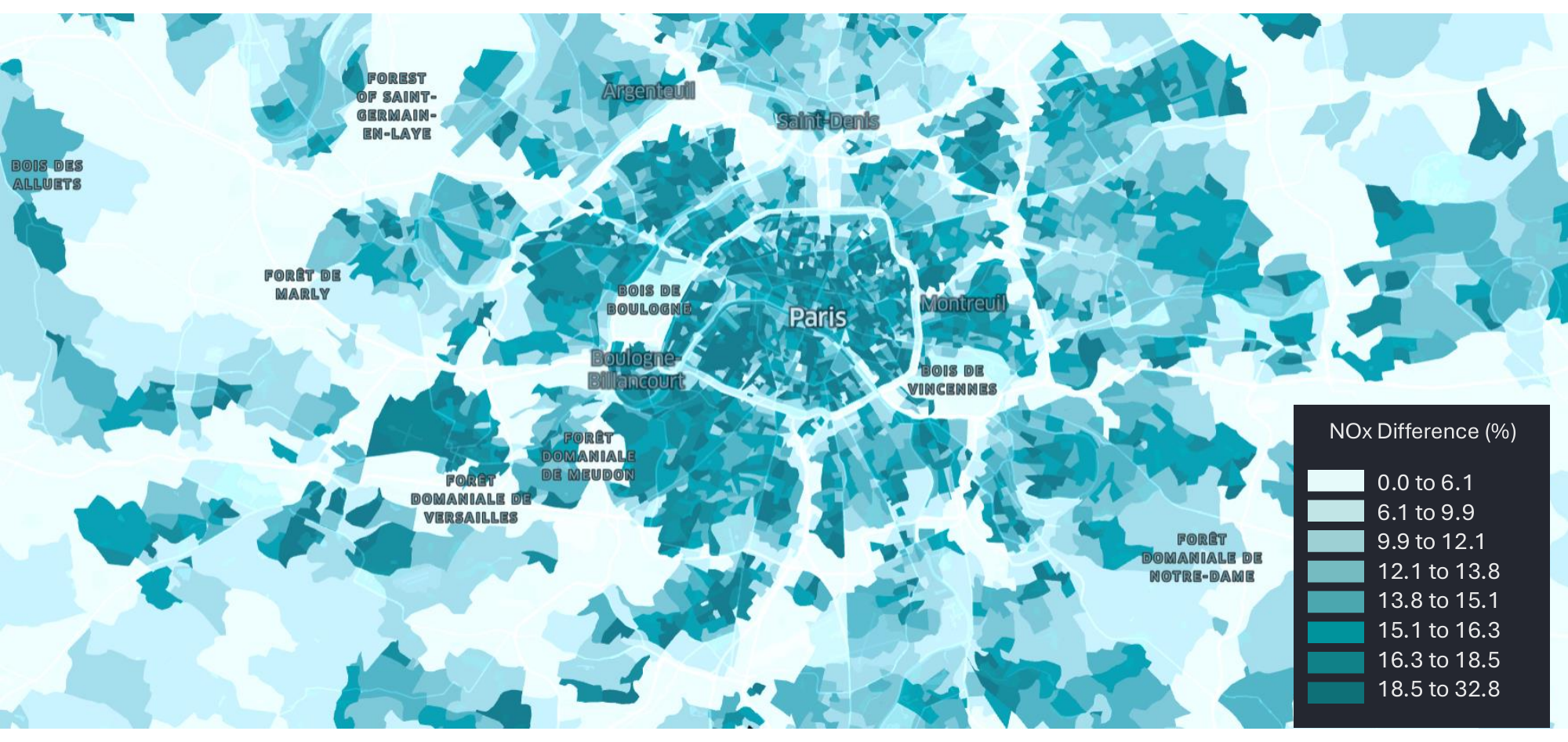}};
        \begin{scope}[shift={(image.south west)}, x={(image.south east)}, y={(image.north west)}]
            \draw[black, fill=none] (0.,0.) rectangle (1.0,1.0);
        \end{scope}
    \end{tikzpicture}
    \caption{Percentage difference in daily NOx emissions at the IRIS level: Hot vs. cold start (at 20°C)}
    \label{fig:temperatureDiff}
\end{figure}

\autoref{fig:temperatureDiff} shows the percentage difference in daily NOx emissions between hot and cold start conditions at 20°C, aggregated at the IRIS level. The results indicate that cold starts have a negligible impact on highway zones but substantially increase emissions in densely populated areas, such as central and greater Paris. This pattern occurs because excess emissions from cold starts are concentrated in the first few kilometers of a trip, typically on local roads characterized by frequent stops and low speeds. By the time the vehicle reaches the highway, it is usually fully or nearly warmed up.

\subsection{Computational Efficiency}

One of the main advantages of DEP is its high computational efficiency. Compared to the SPG-M workflow, DEP takes only 2.2\% of the time required for SPG-M emission estimation (DEP: 1 h 43 min for the 5\% sample IdF scenario on an NVIDIA Quadro P4000 GPU; SPG-M: 3 days 4 h 37 min on CPU, as the emission model is CPU-based). This efficiency enables the rapid generation of detailed spatiotemporal emission maps across a large number of prospective scenarios, a resolution generally not achievable with traditional macroscopic approaches.

\section{Conclusion}\label{sec:conclusion}

This study introduced a novel mesoscopic emission modeling framework that integrates MATSim with DEP, a deep learning-based emission model trained on a synthetic emission dataset. The framework combines the scalability of mesoscopic traffic simulation with the predictive accuracy of deep learning, enabling high-resolution estimation of NOx and CO\textsubscript{2} emissions.

DEP’s sequential architecture, based on GRU networks and designed to process ordered link-level inputs, captures temporal dependencies within trips. It effectively captures acceleration and deceleration patterns as well as cold-start effects, which are key contributors to local emission peaks. This allows for the identification of urban emission hotspots and supports the evaluation of targeted mitigation strategies. The model demonstrates strong predictive performance at both the link and trip levels, as indicated by high $R^2$ values.

Compared to traditional macroscopic models such as COPERT and HBEFA, the proposed approach achieves higher accuracy and reveals that these models tend to underestimate NOx emissions when compared to remote sensing data. When benchmarked against SPG-M, a hybrid mesoscopic–microscopic model, DEP maintains comparable accuracy at global and local scales, while significantly reducing computational time, making it well suited to large-scale policy assessments and scenario analyses.

{Beyond the specific application to the Île-de-France region, this study demonstrates that deep learning-based mesoscopic emission frameworks are feasible and can produce results competitive with far more computationally expensive microscopic approaches, opening the door to scalable, high-resolution emission modeling without reliance on pre-computed emission factor databases.}

However, some limitations persist. Model accuracy depends on the quality of the speed data, which necessitated the use of a speed filter when integrated with MATSim, as MATSim does not model speed transition between links. Furthermore, discrepancies with real-world PEMS measurements, particularly for NOx emissions from Euro 6 vehicles, point to the need for more comprehensive empirical datasets, better calibration, and the inclusion of a wider range of vehicle-specific parameters.

In conclusion, deep learning-enhanced mesoscopic modeling strikes a promising balance between accuracy and computational efficiency. This study demonstrates the potential of such hybrid approaches for scalable and realistic emission estimation. Future work should focus on expanding training datasets, integrating real-world data for fine-tuning, and developing models that are more robust or self-correcting in response to input variability. These improvements will have a major impact on data-driven policies to reduce traffic-related air pollution.

\newpage
{\scriptsize 
\renewcommand{\bibname}{References}
\setlength{\bibsep}{3pt plus 0.3ex}
\bibliographystyle{elsarticle-harv}
\bibliography{references} 
}

\end{document}